# PeaceGAN: A GAN-based Multi-Task Learning Method for SAR Target Image Generation with a Pose Estimator and an Auxiliary Classifier

Jihyong Oh          Munchurl Kim

Korea Advanced Institute of Science and Technology

{jhoh94, mkimee}@kaist.ac.kr

*Abstract*—Although Generative Adversarial Networks (GANs) are successfully applied to diverse fields, training GANs on synthetic aperture radar (SAR) data is a challenging task mostly due to speckle noise. On the one hands, in a learning perspective of human's perception, it is natural to learn a task by using various information from multiple sources. However, in the previous GAN works on SAR target image generation, the information on target classes has only been used. Due to the backscattering characteristics of SAR image signals, the shapes and structures of SAR target images are strongly dependent on their pose angles. Nevertheless, the pose angle information has not been incorporated into such generative models for SAR target images. In this paper, we firstly propose a novel GAN-based multi-task learning (MTL) method for SAR target image generation, called PeaceGAN that uses both pose angle and target class information, which makes it possible to produce SAR target images of desired target classes at intended pose angles. For this, the PeaceGAN has two additional structures, a pose estimator and an auxiliary classifier, at the side of its discriminator to combine the pose and class information more efficiently. In addition, the PeaceGAN is jointly learned in an end-to-end manner as MTL with both pose angle and target class information, thus enhancing the diversity and quality of generated SAR target images The extensive experiments show that taking an advantage of both pose angle and target class learning by the proposed pose estimator and auxiliary classifier can help the PeaceGAN's generator effectively learn the distributions of SAR target images in the MTL framework, so that it can better generate the SAR target images more flexibly and faithfully at intended pose angles for desired target classes compared to the recent state-of-the-art methods.

## I. Introduction

SYNTHETIC aperture radar (SAR) is commonly utilized for surveillance systems [1-4]. Since SAR has a compelling characteristic of a penetration, SAR images can be easily obtained regardless of any weather condition, whether a time is night or daytime and a weather is sunny or cloudy, unlike an optical remote sensing. On the other hand, SAR images generally have serious speckle noise all over the image due to the backscattering of electromagnetic waves [48], so it makes both human and machine-learning algorithms hard to interpret semantic features of the SAR images [5, 6]. However, due to the recent advent of deep learning methods, deep convolutional neural networks (CNNs) have been widely used for many SAR tasks such as recognition of SAR targets [10-13, 58] and optical image classification [31]. However, it is hard to train the CNNs for SAR-related tasks due to the lack of available SAR images that should be obtained costly by radar attached to air vehicles and labeled manually with considerable time consumption [10, 40]. For this reason, there is a need for generative models that can generate abundant SAR data such as "Big Data" for diverse SAR tasks.

Following the first proposal of the Generative Adversarial Networks (GANs) [32], many variants of GANs have shown generative power of GANs for both natural and synthetic images by attaining the similar attributes of their respective original data set [20, 22, 27, 28, 33, 52-57]. Therefore, GAN-based modeling has been one of the most popular generative models [33]. GANs are usually composed of two elements: a generator ($G$) that aims to learn a mapping function from some probabilistic input distributions to a real data distribution; and a discriminator ($D$) that distinguishes whether an input sample comes from the real data distribution or a generative one [32]. At the end, ideally after both players (the generator and discriminator) minimizing their own losses, the training stage approaches to the Nash equilibrium where the generator can produce the images that can hardly be distinguishable by the discriminator whether the samples are real or fake. It is also known that the Nash equilibrium of GANs can be reached by minimizing a statistical divergence between the real data distribution and the generative one as well [32, 35, 37]. After a deep convolutional GAN (DCGAN) was first successfully trained such that both the discriminator and the generator are trained in a well stabilized manner by several deep learning methods, most GANs have begun to adopt the DCGAN architecture as their base structures [34].

Similar to other fields, GANs also have been applied to SAR-related tasks recently. For example, Guo *et al.* first utilized GANs for generating SAR target images based on the basic DCGAN, but they reported that the quality of generated images from a generator were often deteriorated by the so-called *mode collapse* that frequently occurred due to the speckle noise [38]. To prevent those phenomena, they added a pre-processing step, called clutter normalization, for reducing an influence of clutter's speckle noise to boost target recognition performance. On the other hand, Zheng *et al*. utilized multiple discriminators-based GAN model (MGAN) to generate SAR target images without any additional pre-processing by incorporating a semi-supervised learning method [39]. They realized that inherent target-class features from unlabeled SAR target images and label smoothing regularization (LSR) would help improve the generating stability and the quality of generated SAR target images.

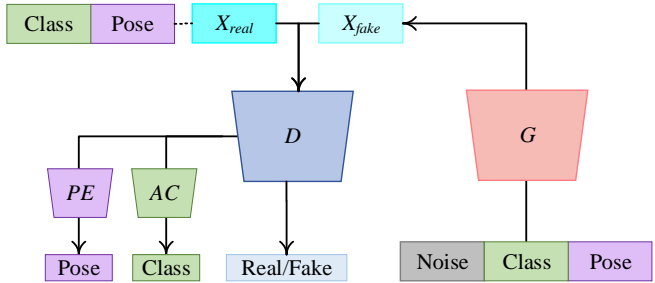

Fig. 1. Overall architecture of our proposed PeaceGAN. The model is composed of four components: a generator (*G*), a discriminator (*D*), pose estimator (*PE*) and auxiliary classifier (*AC*). After a training finishes, the generator can produce abundant SAR target images with intended target classes at desired pose angles.

On the other hand, either pose angle estimation or regression of objects via both CNNs and GANs have been treated as an important auxiliary task to boost the main tasks of classification and generation networks for non-SAR data [41-44]. However, research related to SAR tasks tend to only focus on an intensity information of SAR images and have not taken any pose angle information into account for their main tasks.

Multi-task learning (MTL) that optimizes several loss functions jointly to get information from multiple tasks is well known to be effective in performing its main task in contrast to a single task learning by benefiting from a regularization effect on the usage of multiple objective functions [25, 26]. For example, training a model to predict latent variables in auxiliary classifiers can be hint for a given main task [26]. In addition, MTL makes the model to improve a generalization by collecting joint features from related tasks [25].

To tackle all above issues synthetically, we first present a GAN-based MTL method, called PeaceGAN, for SAR target image generation that can estimate their pose angles by its pose estimator and can learn their target-class features by its auxiliary classifier simultaneously, both of which are located at the side of the discriminator part. Fig. 1 shows the overall structure of our proposed PeaceGAN. To the best of our knowledge, the PeaceGAN is the first approach to jointly learn both the pose angle and the target class information at the same time so that the main task of SAR target image generation via GAN training becomes more stable, thus improving the quality of the resultant generated SAR target images. The contributions of our works are summarized as followings:

- We first propose a novel GAN-based generative model, called PeaceGAN, that is jointly trained in an end-to-end manner with multi-task learning of estimating both pose angle and target class information of SAR target images. The proposed PeaceGAN explicitly disentangles the both pose angles and target classes in learning the distributions of SAR target images, which leads to the enlarged diversity and the increased quality of generated SAR target images.
- Besides, we propose three indirect evaluation methods for SAR target image generation. Based on these, we can measure the quality of generated SAR target images and certify both the generator's ability and the results of GAN training as well as the visually perceived quality.
- Finally, the proposed PeaceGAN can generate SAR target images with both desired pose angles and target classes compared to the recent state-of-the-art methods.

The remainder of this paper is organized as follows: Section II briefly reviews related works on GAN-based methods for SAR target image generation, multi-task learning for GANs, and evaluation metrics for GANs; Section III introduces the proposed GAN-based MTL method for SAR target image generation, called PeaceGAN; Section IV describes the proposed indirect evaluation methods for adequacy and quality of the generated SAR target images; Section V shows experimental results and analyses to demonstrate the effectiveness of the proposed PeaceGAN for SAR target image generation; and Section VI concludes this paper.

## II. RELATED WORKS

### A. GAN-based methods for SAR target image generation

Since training the GANs is generally to solve a minmax problem on the parameters of deep neural networks, it is severely hard in practice for stable learning so that it often fails to appropriately train the GAN-based networks by facing with *mode collapse* problems [30, 37, 47]. Besides, both the standard training methods and the structures of GANs have no significant advantages neither in SAR image classification nor generation [39]. The main reason is the characteristic of SAR images that have the highest spatial resolution collected by X-band SAR with a resolution of 0.3m × 0.3m far lower than those of the optical images. The lower resolution makes both networks and human experience a subtler distinction between different target classes and pose angles of SAR target images. Besides, speckle noise from the interference of reflected wave at the transducer aperture even makes it harder the GANs to be stably trained.

Zheng *et al*. focused on improving SAR target recognition performance of CNN-based framework with semi-supervised GAN learning using multiple discriminators (MGAN) and LSR [39]. The one generator is trained against the feedback aggregated from all the multi-discriminators, which leads the GANs training framework to be more stable. The MGAN also adopts LSR on unlabeled images from the generator in the semi-supervised manner to reduce the confidence of the CNN classifier to get effective supplement, but the multiple discriminators have to be employed with high computation complexities. Besides, the MGAN cannot even directly fuse pose angle information of SAR images into GAN training. Therefore, the MGAN cannot generate SAR target images at desired pose angles of the intended target classes. However, since our proposed PeaceGAN can disentangle both the pose angle and target class information simultaneously, it can generate SAR target images at any given pose angle in a controlled manner, which can provide a high flexibility of generating the SAR target images.

*B. Multi-task learning for GAN-based image synthesis*

The MTL for deep learning is motivated from a biological view of human learning. That is, the human naturally learns a main task by utilizing related tasks as additional information. The harder the main task is, the greater importance of MTL is, in improving both the performance and the generalization of networks [25]. Here, the GANs training is also complex and hard in practice. Tran *et al.* tried to apply MTL into a GAN training framework for an image synthesis and proposed a disentangled representation GAN, called DRGAN that adopts an encoder-decoder structure for its generator by providing pose codes of face variations to the decoder and forces its discriminator to do pose estimation [44]. The class labeling method of DRGAN is similar to a semi-supervised GAN [46] that has *C*-class nodes for classification and one additional node for the real/fake decision at the output of *D*. However, DRGAN is based on an auto-encoder structure for image synthesis, which is totally different from our PeaceGAN. This is because, while the DRGAN is an image-to-image translation network that takes a face image to be rotated into a desired pose angle for pose-invariant face recognition, our PeaceGAN takes noise input with both an intended pose angle and a target class label, and does not translate but generates a corresponding SAR target image of the class at the desired pose angle.

On the other hand, the pose angle information of SAR images has been utilized as *prior* for target recognition, which helps decrease computation complexity and improve target recognition performance [7, 8, 9, 16, 17, 58]. Besides, the backscattered intensities of the same target class SAR images are differently observed at different pose angles due to the back scattering characteristics of speckles [29, 23, 58]. Therefore, it is essential to consider the pose angle information of SAR images for GAN learning. Motivated from this, our PeaceGAN is proposed which can be effectively trained and generate SAR target images of different classes at intended pose angles. This can be accomplished by training the PeaceGAN for MTL with a GANs loss (adversarial loss), a pose angle loss and a target class loss (cross entropy loss).

*C. Evaluation metrics for GANs*

One of the GAN applications is to generate rich data for data augmentation to train deep neural networks. Therefore, it is essential to evaluate the adequacy of generated data as training and test data for intended applications such as classification (target recognition for SAR images in our case). The Frechet inception distance (FID) is most widely used as a metric to evaluate the generated data during GAN training where intra-class mode dropping can be effectively detected [33]. However, this quantitative metric is well suited for only optical images, not SAR images because it utilizes the feature maps of the pre-trained InceptionNet [47] using optical images [14, 33]. The images generated by trained GANs are often evaluated via a user study by using the Amazon Mechanical Turk (MTurk) [36] which is also not suitable for SAR images because the characteristics of SAR images are very different from those of optical images and their interpretations require the expertise in SAR, which is very costly and time-consuming.

On the other hand, there are indirect evaluation methodologies by using pre-trained deep neural networks, which are effective to check the adequacy of generated data for intended target applications. Choi *et al.* [15] trained a facial expression classifier with the Radboud Faces Database, yielding a near-perfect accuracy of 99.55%. They used the pre-trained classifier for quantitative evalutation of their proposed StarGAN [15]. Intuitively, this kind of the indirect evaluation method can be used to judge the suitability of the generated images for the case of either a lower classification error or a larger accuracy rate, that is, whether they can be considered *realistic* with class-distinguishable features or not. Therefore, we adopt this indirect approach for our quantitative evaluations by pre-training classifiers to check the adequacy of the generated SAR target images by our PeaceGAN.

### III. Proposed PeaceGAN – A GAN-based MTL Method for SAR Target Image Generation

In this section, we explain in details about our proposed GAN-based MTL method for SAR target image generation, called PeaceGAN, with a pose estimator and an auxiliary classifier. The PeaceGAN simultaneously learns both the pose angles by a pose estimator and the target-class features by an auxiliary classifier during training for SAR target image generation. By doing so, the PeaceGAN can successfully generate the SAR target images of various classes with high fidelity at any intended pose angle.

*A. Overall structure of proposed PeaceGAN*

An overall structure of the PeaceGAN is shown in Fig. 2. The proposed PeaceGAN is basically composed of a discriminator and a generator. In addition, the discriminator additionally has two newly introduced structures: a pose estimator and an auxiliary classifier for stable generation of SAR target images. We adopt several guidelines of DCGAN [34] to design our proposed PeaceGAN, which have been popularly applied to most settings of GAN training [34]. The architecture of our PeaceGAN is partially based on DCGAN that: i) replaces all the pooling layers with strided convolutions for *D*; ii) utilizes the batch normalization (BN) [50] in both *D* and *G*; iii) assigns ReLU [18] activation functions to all layers of *G* except for the output layer where *Tanh* is used; and iv) uses Leaky ReLU activation functions [7], called lReLU with a slope size of 0.2 for all layers in the discriminator [34]. In the next sub-sections, we describe all components of the PeaceGAN in detail.

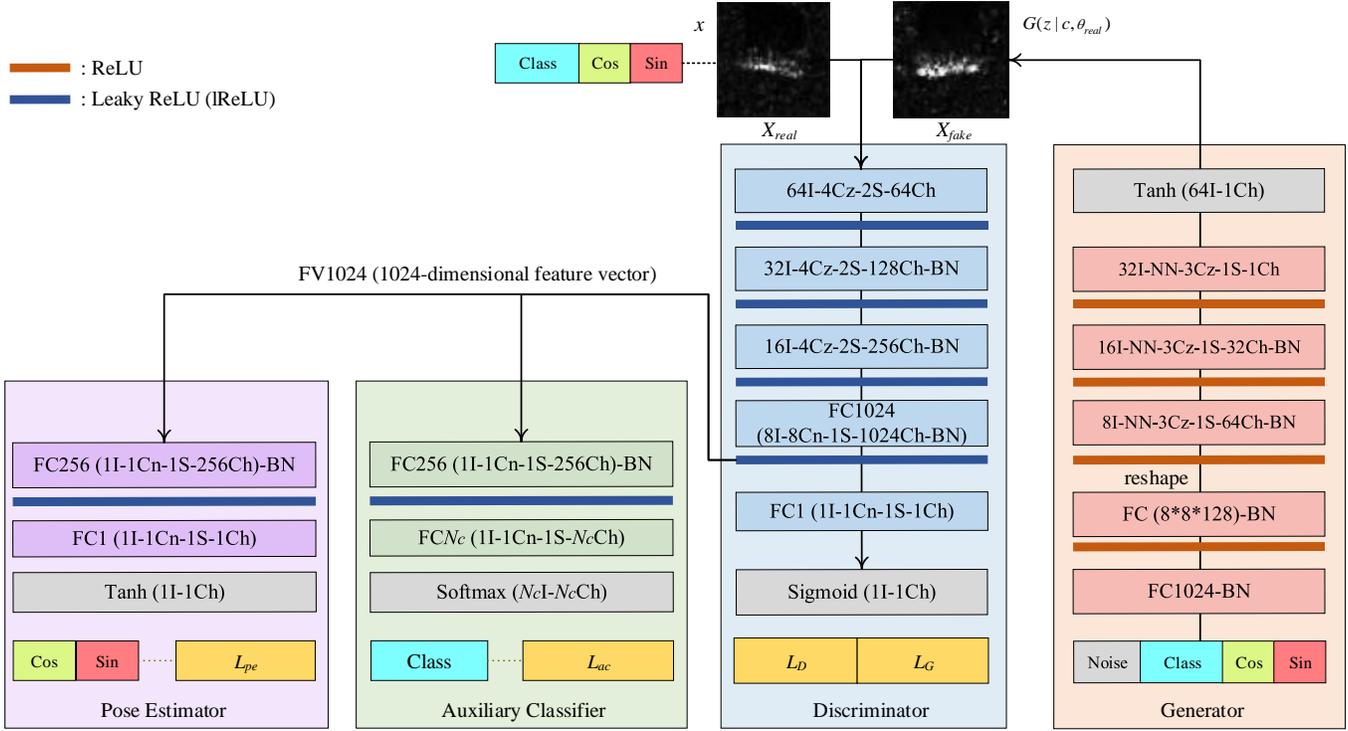

Fig. 2. An overall structure of PeaceGAN: PeaceGAN is basically composed of a discriminator and generator. The generator takes a concatenated input vector of a random noise vector, a target class vector and a pose angle vector. In addition, the discriminator of PeaceGAN additionally has two newly introduced structures: (i) a pose estimator that is trained to estimate the pose angle values of SAR images in terms of both cosine and sine functions, and (ii) an auxiliary classifier that is trained to determine which target classes the SAR images belong to. Note that a convolution layer with $v$I-$w$Cz-$x$S-$y$Ch indicates to have a $v \times v$-sized input, a $w \times w$-sized convolution filter with zero padding, an $x \times x$-sized stride and $y$ output channels. In addition, $w$Cn indicates a $w \times w$-sized convolution filter without zero padding, NN means nearest-neighborhood upsampling with scale 2, BN means batch normalization and FC$y$ means a fully connected layer with output channel size $y$.

*1) Generator of PeaceGAN*

The generator of the PeaceGAN takes a concatenated input vector that is composed of three vectors: a random noise vector, a target class vector and a pose angle vector. The random noise vector $z$ is an $N_z$-dimensional vector (noted as "Noise" in Fig. 2) whose elements are sampled from $P_z(z)$ that is a random uniform distribution in the range of [-1, 1]. The target class vector $c$ is an $N_c$-dimensional one-hot vector with $N_c$ types of target classes, having the value "1" at the corresponding target class location and the values "0" at the others, which is indicated as "Class" at the bottom of the generator in Fig. 2. Lastly, the pose angle vector is a $N_p$ (= 2)-dimensional vector where its first and second elements are the cosine and sine component values for a given SAR target's pose angle $\theta_{real}$ in the range of [0, 360] at which a desired SAR target image is supposed to be generated. The generator utilizes a nearest neighborhood upsampling with scale 2 followed by a 3×3-sized zero padding convolution filter with a 1×1-sized stride indicated as "NN-3Cz-1S", in Fig. 2, called a NNConv to upsample feature maps. The generator has two fully connected (FC) layers and three NNConv's, to finally generate a 64×64 fake SAR target image in the range of [-1, 1] at its output layer.

To sum up, the generator of the PeaceGAN can learn a meaningful mapping function from the synthesized vector with both target class and pose angle of the SAR target image, and generate the SAR target images of high fidelity for desired classes at intended pose angles when the training is successfully done.

*2) Discriminator of PeaceGAN*

The discriminator of the PeaceGAN takes two types of input: one is a real SAR target image and the other is a fake SAR target image generated by the generator, alternatively during the training. Both two types of SAR target images are of a 64×64 size with their values in the range of [-1, 1]. These inputs are sequentially decreased in spatial dimensions by strided convolution filters without any pooling layer and are increased in channel dimensions to extract useful features. The units of lReLU (Leaky ReLU) are used in the discriminator according to the guidelines of DCGAN [34]. The last two layers of the discriminator, denoted as FC1024 and FC1 in Fig. 2, are the fully connected layers with the output channel sizes of 1024 and 1, respectively. Both FC1024 and FC1 are implemented by 8×8 and 1×1 convolution filters, respectively. At the last layer of the discriminator (after "FC1" denoted as in Fig. 2), a sigmoid function yields a scalar output value in the range [0, 1] which is used in the adversarial losses, denoted as "$L_D$" and "$L_G$" in Fig. 2, that will be described in Section III-B in details.

For the purpose of SAR target image generation by jointly learning both pose angles and target classes of SAR target images, we adopt two additional structures: a pose estimator and an auxiliary classifier, both of which are located at the side of the discriminator as shown in Fig. 2. The pose estimator and auxiliary classifier allow for the discriminator not only to distinguish whether the input SAR target images are real or fake, but also to classify the target classes and to estimate pose angles simultaneously via the hard parameter sharing [25] of an MTL framework.

*Pose Estimator (PE)*

As shown in Fig. 2, the pose estimator takes as input a 1024-dimensional feature vector (FV1024) which is the output of the FC1024 layer of the discriminator. The FV1024 input is passed to the FC256 layer followed by the FC1 layer which yields a scalar output, called $T_{pe}$, with the value range of [-1, 1] after *Tanh*. For a pose estimator loss (denoted as $L_{pe}$ in Fig. 2), we first calculate $\theta_{pe} = (T_{pe}+1) \times 180$ that can be viewed as an appropriate pose angle value if the network is well trained with an adequate loss. Then, $\theta_{pe}$ can be converted into the cosine and sine values, respectively, each of which is denoted as "Cos" and "Sin" at the last part (bottom) of the pose estimator in Fig. 2. $L_{pe}$ will be described in Section III-B in details. The newly introduced pose estimator for SAR target image generation encourages both the discriminator and the generator to learn the pose-angle features of SAR target images.

*Auxiliary Classifier (AC)*

Similar to the pose estimator, the auxiliary classifier passes the FV1024 as an input to the FC256 layer followed by the FC$N_c$ layer, and yields an $N_c$-dimensional output, denoted as $S_{ac}$, after a *softmax* activation function [19] for $N_c$ classes of SAR target images. For an auxiliary classifier loss, denoted as $L_{ac}$ in Fig. 2, we use the cross-entropy loss that calculates the difference between the true class probability and the estimated output probability stochastically. $L_{ac}$ will be described in Section III-B in details. The auxiliary classifier for SAR target image generation guides both the discriminator and the generator to learn the scattering characteristics of target-class features of SAR target images.

*Spectral Normalization*

Miyato *et al.* originally applied the spectral normalization (SN) to the discriminator only, for stable training of GAN [27]. After that, Zhang *et al.* argued that applying the SN to the generator also benefits from the GAN training with more stable learning [22]. However, in our SAR target image generation, applying the SN only to the discriminator is almost the same as applying it to both the discriminator and the generator. Therefore, for simplicity, the SN is only applied in the discriminator, including both the pose estimator and the auxiliary classifier with one round of the power iteration method that is known for a sufficient approximation of SN [27].

B. *Combined MTL loss for PeaceGAN*

The PeaceGAN uses the combined loss of the adversarial losses ($L_D$ and $L_G$), the pose estimator loss ($L_{pe}$) and the auxiliary classifier loss ($L_{ac}$) for training. Now we describe these three kinds of losses in detail in the following subsections.

*Adversarial Loss*

In the standard GAN [32], the input of the generator $G$ is generally a random noise vector $z$ with a probability distribution $P_z(z)$, and $G$ generates a fake image $G(z)$, trying to follow the target data distribution $P_{data}$ as closely as possible. The discriminator $D$ alternatively takes two kinds of images: one is a real image $x(=X_{real})$ from $P_{data}(x)$ and the other is a fake image $G(z)(=X_{fake})$. $D$ is trained to distinguish between them well. Finally, both $G$ and $D$ try to solve an adversarial minmax optimization problem with a value function $V(G, D)$ as follows:

$$\min_G \max_D V(G,D) = E_{x \sim P_{data}(x)}[\log D(x)] + E_{z \sim P_z(z)}[\log(1-D(G(z)))] \quad (1)$$

It is known that if both $G$ and $D$ have a sufficient learning capacity, they finally reach the point where neither can improve anymore because $P_{data} = P_{G(z)}$ [32], which is called the Nash equilibrium where $D$ cannot make distinction between real images and fake images. It means that we have $D(x) = D(G(z)) = 0.5$ [32]. In addition, Eq. (1) can be solved in the form of a non-saturating (NS) loss as follows [32]:

$$L_D = -E_{x \sim P_{data}(x)}[\log D(x)] - E_{z \sim P_z(z)}[\log(1-D(G(z)))] \quad (2)$$

$$L_G = -E_{z \sim P_z(z)}[\log D(G(z))] \quad (3)$$

The final output of $D$ is usually a sigmoid function value so that a sigmoid cross-entropy loss is generally utilized.

For our proposed PeaceGAN, since we adopt an MTL framework for GAN training with the auxiliary classifier and the pose estimator, it has two additional components: pose angle value $\theta_{real}$ and target class $c$, concatenated to the noise vector $z$ at the input part of the generator that generates a fake SAR target image $G(z|c, \theta_{real})$. The real images $x$ from $P_{data}(x)$ and fake images $G(z|c, \theta_{real})$ are alternatively fed into the discriminator $D$. Additionally, we adopt the WGAN-GP loss [30] as the adversarial loss for the PeaceGAN for training well, which can be expressed in our case as:

$$L_D = -E_{x \sim P_{data}(x)}[D(x|c, \theta_{real})] + E_{z \sim P_z(z)}[D(G(z|c, \theta_{real}))] + \lambda_{gp} E_{\hat{x} \sim P_{\hat{x}}(\hat{x})}[(\|\nabla_{\hat{x}} D(\hat{x})\|_2 - 1)^2] \quad (4)$$

$$L_G = -E_{z \sim P_z(z)}[D(G(z|c, \theta_{real}))] \quad (5)$$

where $\hat{x}$ is defined as $\hat{x} = \varepsilon x + (1-\varepsilon) G(z|c, \theta_{real})$, $\varepsilon$ is randomly sampled from the uniform distribution, i.e. $\varepsilon \sim U[0,1]$ and $\lambda_{gp}$ is a weight parameter for WGAN-GP [30]. The gradient norm $E_{\hat{x} \sim P_{\hat{x}}(\hat{x})}[(\|\nabla_{\hat{x}} D(\hat{x})\|_2 - 1)^2]$ for the penalty is calculated on a linear interpolation between the pairs of real data points $P_{data}(x)$ and those of the generator distribution $G(z|c, \theta_{real})$ [30]. It should be noted that WGAN-GP utilizes the *critic* directly for losses rather than using the sigmoid cross-entropy loss [37]. The adversarial loss guides the generator of the PeaceGAN to generate fake SAR target images which are not distinguishable from the real ones by the discriminator of the PeaceGAN. Note that we directly use FC1 output (i.e. *critic*), not "Sigmoid" of Fig. 2, for (4).

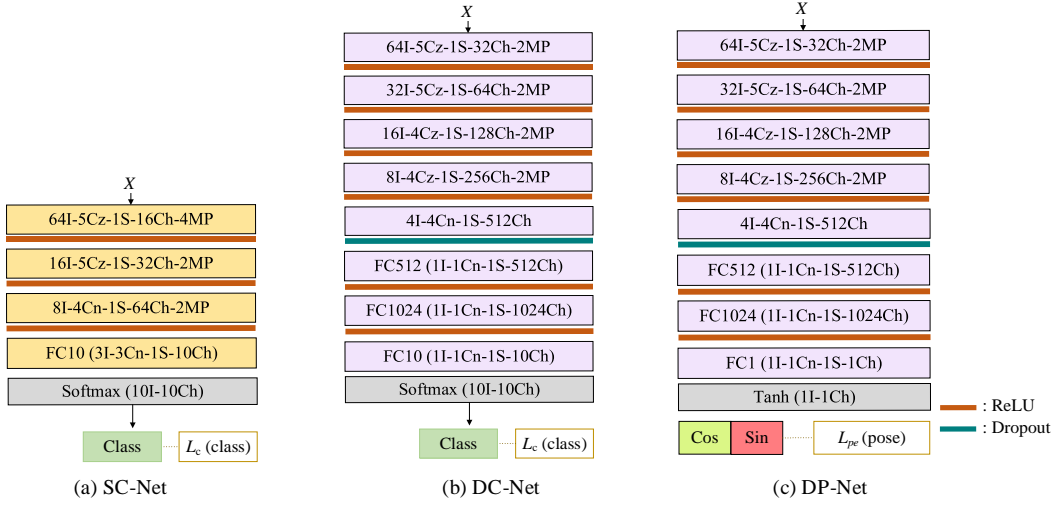

Fig. 3. The three types of network structures used for proposed indirect evaluation methods for SAR target image generation: (a) simple classifier (SC-Net) (b) deep classifier (DC-Net) (c) deep pose-estimator (DP-Net).

*Pose Estimator Loss*

The loss $L_{pe}$ of the pose estimator has two components of $L_{pe,real}$ and $L_{pe,fake}$ for the input data of two types into the discriminator. The two losses are calculated by

$$L_{pe,real} = E_{x \sim P_{data}(x)}[(\cos\theta_{real} - \cos\theta_{pe})^2 \\ + (\sin\theta_{real} - \sin\theta_{pe})^2 \mid x] \quad (6)$$

$$L_{pe,fake} = E_{z \sim P_z(z)}[(\cos\theta_{real} - \cos\theta_{pe})^2 \\ + (\sin\theta_{real} - \sin\theta_{pe})^2 \mid G(z|c, \theta_{real})] \quad (7)$$

$L_{pe,real}$ in (6) guides the discriminator to identify the pose angles of SAR target images as correctly as possible, and $L_{pe,fake}$ in (7) allows for the generator to have an ability to generate SAR target images at the intended pose angles.

*Auxiliary Classifier Loss*

The loss $L_{ac}$ of the auxiliary classifier is calculated by a cross-entropy measure to calculate the difference between estimated target class probabilities $S_{ac}$ and true target-class probabilities $P_c(c)$, which is calculated as

$$L_{ac,real} = E_{x \sim P_{data}(x)}[\log P(S_{ac} = c \mid x)] \quad (8)$$

$$L_{ac,fake} = E_{z \sim P_z(z)}[\log P(S_{ac} = c \mid G(z|c, \theta_{real}))] \quad (9)$$

$L_{ac,real}$ in (8) forces the discriminator to classify the input SAR target images as accurately as possible, while $L_{ac,fake}$ in (9) has the generator to produce the SAR target images of the desired classes as precisely as possible.

*Total Combined Losses*

The total combined losses for the PeaceGAN are denoted as $L_{PeaceGAN,D}$ for its discriminator and $L_{PeaceGAN,G}$ for its generator, respectively, which are given by

$$L_{PeaceGAN,D} = \lambda_{adv}L_D + \lambda_{mtl}(\lambda_{pe}L_{pe,real} + \lambda_{ac}L_{ac,real}) \quad (10)$$

$$L_{PeaceGAN,G} = \lambda_{adv}L_G + \lambda_{mtl}(\lambda_{pe}L_{pe,fake} + \lambda_{ac}L_{ac,fake}) \quad (11)$$

where $\lambda_{adv}$, $\lambda_{pe}$, $\lambda_{ac}$ and $\lambda_{mtl}$ are weight parameters. These two total combined losses are used to train the proposed PeaceGAN in an MTL framework, which allows for effective GAN learning. As a result, if the training of the PeaceGAN is successfully finished, the generator can produce abundant SAR target images of desired classes at any pose angle.

## IV. PROPOSED INDIRECT EVALUATION METHODS FOR ADEQUACY AND QUALITY OF GENERATED SAR TARGET IMAGES

In order to inspect the effectiveness of the PeaceGAN, we introduce new indirect evaluation methods in a perspective of GAN-based SAR target image generation. It is important to evaluate the both the adequacy of the generated SAR target images by the generator as training data and their fidelities of pose-angle features and target-class features. Nevertheless, none of the methods has addressed this issue in quantitative manners. In this paper, we propose three ways of evaluating the generated SAR target images: (i) the *adequacy* as training data for SAR target recognition using a simple classifier, called SC method; (ii) the *quality* of generated SAR target images using an over-fitted deep classifier for real SAR target images of specific target classes, called DC method; and (iii) the *quality* of generated SAR target images using an over-fitted deep pose-estimator for real SAR target images at specific pose angles, called DP method. Fig. 3 shows the three types of network structures used for the indirect evaluation methods. From now on, we assume that GAN's generators have finished to learn the distribution of real SAR target images with a depression angle of 17°, called $D_{17}$.

### A. Indirect Evaluation using SC Method

Fig. 3-(a) shows a simple classifier for the SC method, called

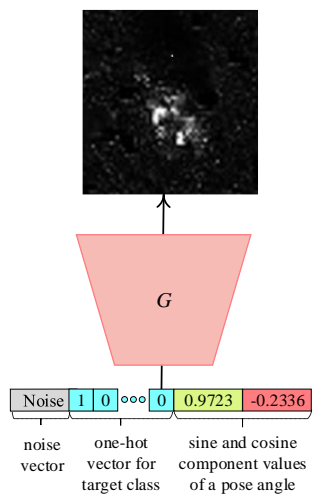

Fig. 4. An example of how to generate the intended class of the SAR image with the desired pose angle by PeaceGAN: The 'BMP2' -class SAR image (first class) with pose angle 346.49° of $D_{17}$.

SC-Net, that has a softmax activation function in the softmax cross-entropy loss as for SAR target recognition cases [13]. In addition, the SC-Net is designed to have a small number of convolution layers where each convolution layer consists of a small number of feature channels to yield a relatively lower target recognition performance because it should not memorize the generated SAR target images during training. Then, the SC-Net trained with the generated SAR target images can fairly be tested for real SAR target images. By doing so, it is possible to make two comparisons: (i) training the SC-Net with generated SAR target images of a depression angle of 17° by a GAN-based 'X' method, called $D_{17,GAN}^{X}$, and testing the trained SC-Net for real SAR target images with a depression angle of 15°, called $D_{15}$; and (ii) training the SC-Net with real SAR target images with a depression angle of 17° ($D_{17}$), and testing the trained SC-Net for $D_{15}$. Note that, by using the trained PeaceGAN, we can generate $D_{17,GAN}^{Peace}$ with the same number of SAR target images per class at the same pose angles corresponding to each real SAR image in $D_{17}$ because PeaceGAN's generator can produce the SAR target images of an intended class at any desired pose angle by feeding into the generator the combined input vector of the corresponding cosine and sine values of a given pose angle and the one-hot vector for a target class, as shown in Fig. 4. In addition, $D_{15}$ is generally used as test data for SAR target recognition when $D_{17}$ is used as train data [13]. More information on the $D_{17}$ and $D_{15}$ for our experiments will be described in Section V-A in detail.

### B. Indirect Evaluation using DC Method

Fig. 3-(b) shows a deep classifier for the DC method, called DC-Net. Unlike the SC method, the DC-Net is pre-trained to indirectly evaluate (inspect) the *quality* of $D_{17,GAN}^{X}$ in the perspectives of target-class features. First, the DC-Net is intentionally trained with a training set of 10,000 samples augmented from $D_{17}$ to yield 100% target recognition performance for $D_{17}$. Here, the DC-Net has a deep structure with a large number of convolution layers, each of which has a large number of feature channels, to be over-fitted for $D_{17}$.

Therefore, in a testing phase, a higher target recognition performance of the pre-trained DC-Net for $D_{17,GAN}^{X}$ implies that the better quality of $D_{17,GAN}^{X}$ is generated by the GAN's generator, and the generator is well trained in the perspective of target-class features. We pre-trained the DC-Net five times with a random weight initialization each time to have the 100% classification rates for all experiments. So, the quality of $D_{17,GAN}^{X}$ is measured as the average of the resulting five target classification rates tested by the five pre-trained DC-Nets in the perspective of target-class features.

### C. Indirect Evaluation using DP Method

Fig. 3-(c) shows a deep pose-estimator for the DP method, called DP-Net. Also, the DP-Net is designed almost in the same manner as the DC-Net. The only difference is the usage of *Tanh* at the output where the DP-Net is trained by the same square difference loss in (6). Also, using the same training set of 10,000 samples augmented from $D_{17}$, we have pre-trained the DP-Net five times with a random weight initialization each time for pose angle estimation where the DP-Net is trained to be over-fitted to the pose-angle features of $D_{17}$ each time. It should be noted that, unlike the DC-Net, it is hard to train the DP-Net to be over-fitted to perfectly predict the pose angles of the samples in $D_{17}$. This is because the pose angles of the SAR target images have continuous values in the range of [0°, 360°). From the five-times training of over-fitting, the DP-Net yielded the pose angle estimation performance in terms of the two averages of the resulting five MAD and STD values with 2.22° and 3.28°, respectively, for $D_{17}$ where the MAD and STD indicate the mean absolute difference and the standard deviation difference between the true pose angles ($\theta_{real}$) and their estimated ones ($\theta_{pe}$), respectively. Therefore, when measuring the quality of $D_{17,GAN}^{X}$ in the perspective of pose-angle features, smaller MAD and STD values measured by the pre-trained DP-Net for $D_{17,GAN}^{X}$ imply that a better quality of $D_{17,GAN}^{X}$ is generated and the corresponding generator is also trained well in the perspective of pose-angle features.

The SC-Net, DC-Net and DP-Net in Fig. 3 are trained with the same hyper-parameters. All weights of the convolution filters are initialized by Xavier initializer [49]. The mini-batch size is set to 100, the weight decay coefficient for L2 loss is 0.004, the total epoch is 50, the initial learning rate is set to 0.01 and is multiplied by 0.1 cumulatively at the 10-, 25- and 40-th epochs. In addition, the mini-batch stochastic gradient descent method is utilized with a momentum parameter value of 0.9.

## V. EXPERIMENTS RESULTS

### A. Experiments settings

#### MSTAR Dataset

The Moving and Stationary Target Acquisition and Recognition (MSTAR) public dataset is utilized for experiments of the SAR target image generation. The MSTAR was gathered by the Sandia National Laboratory (SNL) SAR sensor platform that was supported by Defense Advanced

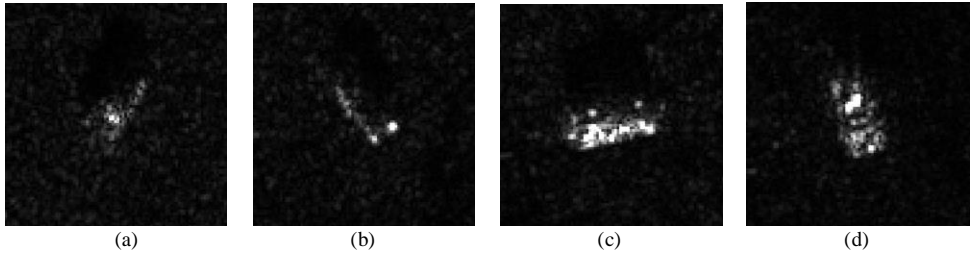

(a) (b) (c) (d)

Fig. 5. Examples of MSTAR dataset: (a) 'BMP2'-class with 26.49 pose angles and 17 depression angles, (b) '2S1-class with 143.33 pose angles and 17 depression angles, (c) 'T62-class with 256.52 pose angles and 17 depression angles, (d) 'ZSU234'-class with 347.99 pose angles and 15 depression angles.

TABLE I
THE INFORMATION OF MSTAR DATA FOR EXPERIMENTS.

| Target names (classes) | Serial # | For GANs, DC and DP (SAR image size: 64×64) | | Only used for SC (SAR image size: 64×64) | |
|---|---|---|---|---|---|
| | | Depression angles | # of data (total:2747) | Depression angles | # of data (total:2425) |
| BMP2 | 9563 | 17 | 233 | 15 | 195 |
| BTR70 | C71 | 17 | 233 | 15 | 196 |
| T72 | 132 | 17 | 232 | 15 | 196 |
| BTR60 | K10 | 17 | 256 | 15 | 195 |
| 2S1 | B01 | 17 | 299 | 15 | 274 |
| BRDM2 | E71 | 17 | 298 | 15 | 274 |
| D7 | 92V | 17 | 299 | 15 | 274 |
| T62 | A51 | 17 | 299 | 15 | 273 |
| ZIL131 | E12 | 17 | 299 | 15 | 274 |
| ZSU234 | D08 | 17 | 299 | 15 | 274 |

Research Projects Agency (DARPA) and Air Force Research Laboratory (AFRL) [21]. Each MSTAR data includes information about azimuth angles, depression angles, versions, configurations, and target classes, etc. MSTAR dataset is a benchmark for experiments related to various problems handling SAR images [10-13, 38, 39].

Fig. 5 shows some SAR target images of four classes having different pose angles of two different depression angles taken from the MSTAR data. As shown in Fig. 5, it is hard to infer either the exact target classes or the corresponding pose angles of the SAR target images due to both their speckle noise and low spatial resolutions. For experiments, we use a data set of the central-cropped 64×64-sized 2,747 SAR target images with a depression angle of 17° for $D_{17}$ from MSTAR data to train GANs. The size of 64×64 is known to be sufficient to contain both the clutters and the targets adequately for SAR target image generation [38]. $D_{15}$ also consists of the 64×64-sized 2,425 SAR target images of MSTAR data obtained under a depression angle of 15°. Table I summarizes the data set used for our experiments.

Since there are limited numbers of available MSTAR's SAR target images for GAN training, we utilized data augmentation methods of [12] and [13]: (i) *rotation*, (ii) *pose synthesis* and (iii) *speckle noising* to prevent from the *mode collapse* and a memorization problem of the GAN generators, which are the severe issues of GAN training. *Rotation* data augmentation randomly rotates SAR target images to produce new SAR target images within a small 15-degree angle to preserve the backscattering characteristics [12, 13]. *Pose synthesis* data augmentation combines neighboring two SAR images, which have similar pose angles, to generate new SAR target images by a weighted sum [12]. *Speckle noising* data augmentation generates new data SAR images by newly adding speckle noises sampled from an exponential distribution [12]. To conduct experiments at sufficient performance levels for better comparisons, we produced 5,000 SAR target images for each of the 10 classes based on the above data augmentation methods. As a result, our training data set has 64×64-sized 50,000 SAR target images with the 17° depression angle for all 10 classes.

*Implementation of PeaceGAN*

We use two Adam optimizers [24] to minimize $L_{PeaceGAN, G}$ and $L_{PeaceGAN, D}$ with learning rates of $lr_g$ and $lr_d$, respectively. The initial value of $lr_d$ is set to 0.0005 that starts to decay linearly to zero from the 40-th epoch to the last 60-th epoch. In addition, we found that a relationship of $lr_g = 5 \times lr_d$ is appropriate for stable training in our experiments. The generator and the discriminator with the two additional structures are trained alternatively by the two corresponding Adam optimizers. The momentum term $\beta_1$ of Adam optimizers is set to 0.5 that is widely known to make the GAN learning more stable [34], and $\beta_2$ is set to 0.999 as a default setting [24]. All weights of the convolution filters are initialized by random normal distribution with a zero mean and a standard deviation of 0.02. The mini-batch size is 25. The dimension of the input noise vector is $N_z = 64$, the number of SAR target image classes is $N_c = 10$, and $N_p = 2$ since the pose angle input consists of two sine and cosine angle components. For the weight parameters for losses, we empirically set $\lambda_{gp} = 5$, $\lambda_{adv} = 1$, $\lambda_{pe} = 0.2$, $\lambda_{ac} = 0.8$ and $\lambda_{mtl} = 1$. The size of all SAR target images for experiments is 64×64.

B. *Evaluations on Generated SAR Target Images by the proposed PeaceGAN*

*Subjective Comparison between Generated and Real SAR Target Images*

Fig. 6 shows some generated SAR target images in $D_{17,GAN}^{Peace}$ generated by the PeaceGAN and their corresponding real SAR target images of a same target class at the same pose angles in $D_{17}$. As shown in Fig. 6, it can be noted that the PeaceGAN can produce SAR target images visually very similar to the actual SAR target images. The generated SAR target images contain similar speckle noises to those of the real ones and similar shadows around the targets.

*Analysis for the performances of three PeaceGAN variants using SC and DC*

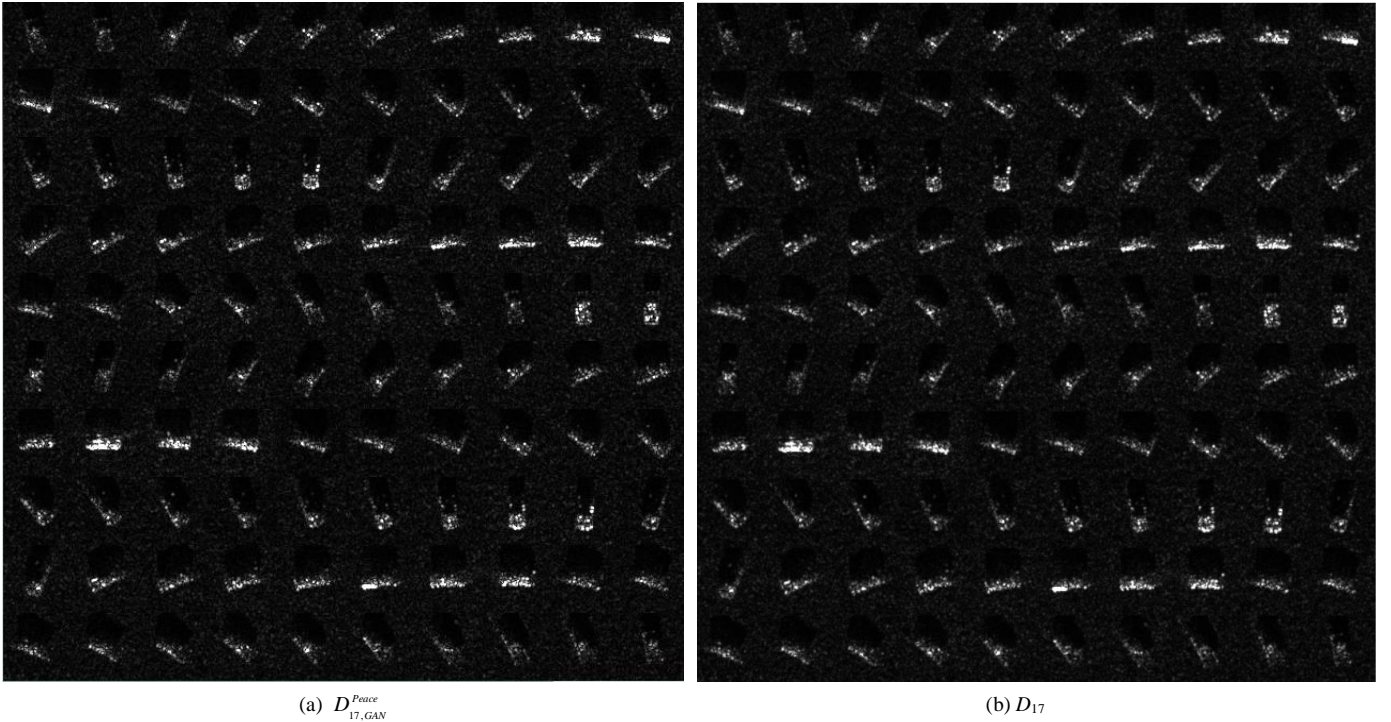

(a) $D_{17,GAN}^{Peace}$      (b) $D_{17}$

Fig. 6. The comparison between $D_{17,GAN}^{Peace}$ and $D_{17}$: (a) First 100 'BMP2'-class SAR images generated by PeaceGAN at pose angle increments with 5°, (b) First original 100 'BMP2'-class SAR images from $D_{17}$ at pose angle increments with 5°.

TABLE II
TARGET RECOGNITION PERFORMANCE (%) FOR PEACEGAN VARIANTS INDIRECTLY EVALUATED BY THE SC AND DC METHODS

| Structural Variants | with SN | | without SN | |
|---|---|---|---|---|
| | SC | DC | SC | DC |
| **PeaceGAN with NNConv & $CS_g$** | **81.20** | **99.42** | 78.93 | 98.46 |
| PeaceGAN with USConv & $CS_g$ | 77.61 | 98.04 | 76.25 | 98.01 |
| PeaceGAN with NNConv & $DN_g$ | 76.21 | 96.70 | 74.56 | 96.07 |

We compare three PeaceGAN variants with respect to their structures using the SC and DC (methods) from a structural point of view. Table II shows the target recognition performance (%) for the three PeaceGAN variants indirectly evaluated by the SC and DC. For this, each $D_{17,GAN}^{Peace}$ generated by PeaceGAN variant is utilized to train the SC-Net three times with a random weight initialization to obtain stable average results tested with $D_{15}$, according to the SC described in Section IV-A, and the each $D_{17,GAN}^{Peace}$ is tested by the five pre-trained DC-Nets to yield an average target recognition performance, according to the DC described in Section IV-B. In Table II, the 'PeaceGAN with NNConv & $CS_g$', which is our baseline structure shown in Fig. 2, indicates that (i) its generator uses the nearest neighborhood upsampling method as described in Section III-A-1); and (ii) the pose angle input to the generator consists of separate cosine and sine component values, called $CS_g$, trained with (6) and (7) for the pose estimation losses. Instead of the NNConv, the 'PeaceGAN with USConv & $CS_g$' utilizes the fractionally-strided convolution filters for two-times upsampling, called USConv, used in the DCGAN's generator [34]. Finally, instead of using the $CS_g$ as a pose angle input, the 'PeaceGAN with NNConv & $DN_g$' replaces the $CS_g$ with two same values, each of which is the directly normalized pose angle by $\theta_{norm} = (\theta_{real} - 180)/180$ into the range of [-1, 1] to keep $N_p = 2$ for a fair comparison to $CS_g$, denoted as $DN_g$, and this variant is trained with following (12) and (13) for the pose estimation losses instead of (6) and (7) to keep a scale of losses for the fair comparison:

$$L_{pe,real} = E_{x \sim P_{data}(x)}[2 \times (\theta_{norm} - \theta_{pe})^2 \mid x] \quad (12)$$

$$L_{pe,fake} = E_{z \sim P_z(z)}[2 \times (\theta_{norm} - \theta_{pe})^2 \mid G(z \mid c, \theta_{norm})] \quad (13)$$

It is also should be noted that both (12) and (13) directly calculate mean squares error between $\theta_{norm}$ and $\theta_{pe}$, not utilizing both cosine and sine functions. From the results in Table II, it can be noted that NNConv is more effective than USConv, and the usage of $CS_g$ in the form of separate cosine and sine components trained with (6) and (7) loss functions also help the generator effectively learn the target-class features with respect to the target's pose angles better than the usage of $DN_g$. It is also worthwhile to mention that applying the spectral normalization (SN) to the discriminator of the PeaceGAN is also effective for GAN-based SAR target image generation, as shown in Table II.

Finally, in order to see the adequacy of the generated SAR target image set ($D_{17,GAN}^{Peace}$) as a training dataset compared to its corresponding original SAR target dataset ($D_{17}$), the SC-Net was also trained with $D_{17}$ and yielded 83.71% target recognition

TABLE III

CONFUSION MATRIX FOR TARGET RECOGNITION PERFORMANCE (%) ON $D^{Peace}_{17,GAN}$ TESTED BY FIVE PRE-TRAINED DC-NETS.

| Classes | BMP2 | BTR70 | T72 | BTR60 | 2S1 | BRDM2 | D7 | T62 | ZIL131 | ZSU234 | Recognition rates (%)/Std. |
|---|---|---|---|---|---|---|---|---|---|---|---|
| BMP2 | 230.6 | 1 | 0 | 1.2 | 0.2 | 0 | 0 | 0 | 0 | 0 | 98.97 /0.210 |
| BTR70 | 1 | 229.4 | 0 | 2.6 | 0 | 0 | 0 | 0 | 0 | 0 | 98.45/0.210 |
| T72 | 0 | 1 | 231 | 0 | 0 | 0 | 0 | 0 | 0 | 0 | 99.57/0.000 |
| BTR60 | 0 | 4.2 | 0 | 251.8 | 0 | 0 | 0 | 0 | 0 | 0 | 98.36/0.156 |
| 2S1 | 0 | 1 | 0 | 0 | 296.6 | 0 | 0 | 1.4 | 0 | 0 | 99.20/0.164 |
| BRDM2 | 0.2 | 0 | 0 | 1 | 0 | 296.8 | 0 | 0 | 0 | 0 | 99.60/0.134 |
| D7 | 0 | 0 | 0 | 0 | 0 | 0 | 299 | 0 | 0 | 0 | 100.0/0.000 |
| T62 | 0 | 0 | 1 | 0.2 | 0 | 0 | 0 | 297.8 | 0 | 0 | 99.60/0.134 |
| ZIL131 | 0 | 0 | 0 | 0 | 0 | 0 | 0 | 0 | 299 | 0 | 100.0/0.000 |
| ZSU234 | 0 | 0 | 0 | 0 | 0 | 0 | 0 | 0 | 0 | 299 | 100.0/0.000 |
| Averages | | | | | | | | | | | **99.42/0.025** |

TABLE IV

POSE ESTIMATION PERFORMANCE OF THE PRE-TRAINED DP-NET ON $D_{17}$ AND $D^{Peace}_{17,GAN}$.

| Classes | $D_{17}$ | | $D^{Peace}_{17,GAN}$ | |
|---|---|---|---|---|
| | MAD | STD | MAD | STD |
| BMP2 | 2.14 | 3.00 | 7.03 | 11.49 |
| BTR70 | 2.31 | 3.67 | 8.48 | 12.04 |
| T72 | 2.61 | 3.89 | 6.71 | 10.44 |
| BTR60 | 2.16 | 3.47 | 6.71 | 9.55 |
| 2S1 | 1.90 | 2.75 | 6.84 | 10.29 |
| BRDM2 | 2.12 | 2.99 | 6.77 | 10.73 |
| D7 | 2.37 | 3.83 | 7.45 | 10.38 |
| T62 | 2.11 | 2.66 | 6.39 | 10.27 |
| ZIL131 | 2.16 | 2.97 | 5.83 | 8.49 |
| ZSU234 | 2.37 | 3.58 | 6.10 | 8.99 |
| Averages | **2.22** | **3.28** | **6.83** | **10.27** |

performance when it was tested with $D_{15}$. On the other hand, when the SC-Net was trained with $D^{Peace}_{17,GAN}$ and tested with $D_{15}$, the resulting target recognition rate turned out to be 81.20% as shown in Table II, which is very close to 83.71%. From this, it can be noted that the PeaceGAN is capable of generating adequate SAR target images as training data. As mentioned before, the DC-Net was pre-trained in an intended over-fitted manner for $D_{17}$, thus yielding 100% target recognition performance for $D_{17}$ itself. The intention of this over-fitting is that the DC-Net is pre-trained to memorize the target-class features of $D_{17}$. When $D^{Peace}_{17,GAN}$ is tested by the five pre-trained DC-Nets, the averaged target recognition performance turned out to be 99.42% as shown in Table II, which is very close to 100%. For the target recognition performance of 99.42% in Table II, Table III shows in details the averaged per-class target recognition performance (confusion matrix) of the five pre-trained DC-Nets for $D^{Peace}_{17,GAN}$. As shown in Table III, the generated SAR target images of 'D7', 'ZIL131' and 'ZSU234' classes are classified remarkably with 100% accuracy and those of the other classes are also classified nearly close to 100% accuracy. From this, it can also be noted that the PeaceGAN is capable of generating the SAR target images with very similar target-class features of $D_{17}$.

*Indirect Evaluation by DP for Pose-Angle Features of $D^{Peace}_{17,GAN}$*

One of the PeaceGAN's advantages is a capability to generate SAR target images at intended pose angles. So, it is necessary to evaluate whether or not the PeaceGAN can appropriately generate SAR target images in a perspective of pose angles. For this, we use an indirect evaluation method using the DP method as described in Section IV-C. Table IV shows the pose estimation performance of the pre-trained DP-Net on $D_{17}$ and $D^{Peace}_{17,GAN}$. The estimation accuracy values in Table IV are measured in terms of MAD and STD as their averages of five-times tests by the DP-Net with five-times independent training with random weight initialization each time. The DP-Net shows the average 2.22° MAD for $D_{17}$, and the average 6.83° for $D^{Peace}_{17,GAN}$. It can be noted in Fig. 6-(b) that the generated SAR target images at neighboring pose angles with 5° differences are visually very similar. So, the pose-angle features of $D^{Peace}_{17,GAN}$ are sufficient similar to those of $D_{17}$.

### C. Comparison between PeaceGAN, modified ACGAN and modified CGAN

To highlight the novelty of our PeaceGAN in a structural point of view, we compare it with the most relevant GANs, ACGAN [51] and CGAN [41]. The ACGAN is based on the standard GAN and has an additional component of a classifier, which is called the auxiliary classifier [51]. The original ACGAN only utilizes class information that is concatenated to the input vector of its generator and has the same forms of classification losses in (8) and (9) but that is directly connected to the end of its discriminator, not independently such as the structure of the auxiliary classifier of the PeaceGAN. We compare the generation capabilities of SAR target images by the PeaceGAN that utilizes both target class and pose angle information and the ACGAN that only uses target class information. For a fair comparison, we modify the original ACGAN to have the same generator, discriminator and auxiliary classifier of the PeaceGAN, called ACGAN$_m$, which becomes equivalent to a PeaceGAN variant without the pose estimator. For experiments, the parameters of the ACGAN$_m$ are set with the same values as the PeaceGAN, except for the two parameters with $N_z = 66$ (= 64 + 2), $N_c = 10$ and $N_p = 0$ to keep the original 76-dimensional combined input vector, and with $\lambda_{pe} = 0$ and $\lambda_{ac} = 1.0$ (= 0.8 + 0.2) to keep the same scales of gradients. Since the ACGAN$_m$ cannot incorporate the pose angle information directly into SAR target image generation,

## TABLE V
THE COMPARISON BETWEEN $D_{17,GAN}^{Peace}$, $D_{17,GAN}^{AC}$ AND $D_{17,GAN}^{C}$: TARGET RECOGNITION PERFORMANCE (%) BY SC AND DC.

| Structures | SC | DC |
|---|---|---|
| **PeaceGAN** | **81.20** | **99.42** |
| $ACGAN_m$ | 73.03 | 96.23 |
| $CGAN_m$ | 77.53 | 95.63 |

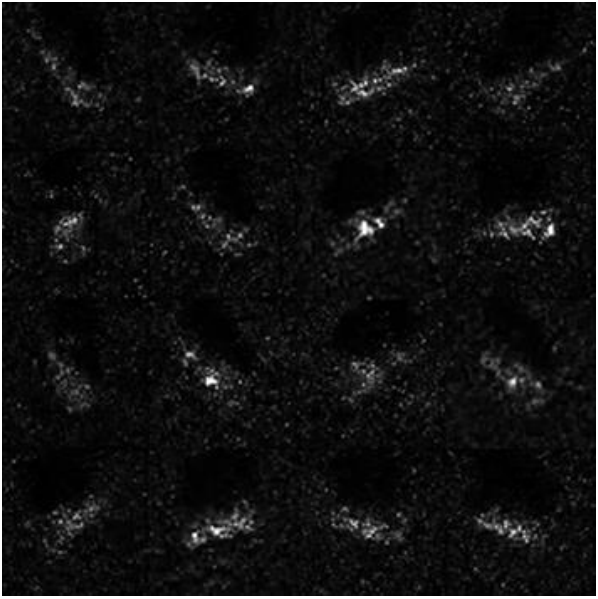

Fig. 7. The 16 'BMP2'-class SAR images generated by $ACGAN_m$. It should be noted that the generator of $ACGAN_m$ generate SAR target images at random pose angles with the same number of SAR target images per each class corresponding to $D_{17}$.

we are only able to constitute the input vector with an one-hot vector of target classes and a random noise vector, where the resulting generated SAR target images are denoted as $D_{17,GAN}^{AC}$ for the depression angle of 17°. So $D_{17,GAN}^{AC}$ becomes to have generated SAR target images at random pose angles with the same number of SAR target images per each class corresponding to $D_{17}$.

On the other hand, we also design a modified CGAN-based network, called $CGAN_m$. For a fair comparison, we firstly modify the original CGAN [41] to have the same generator and discriminator of the PeaceGAN without both pose estimator and auxiliary classifier. Secondly, to follow the definition of CGAN [41], we design a discriminator of $CGAN_m$ additionally having conditioned inputs that composes of the $X_{real}$ or $X_{fake}$ concatenated with a stretched (filling same values in 64×64 spatial dimensions) one-hot target class, a cosine value and a sine value of a pose angle. For experiments, the all parameters of the $CGAN_m$ are set with the same values as the PeaceGAN except for the input channel of the discriminator increased to 13 (= 1+10+2) from 1 and $\lambda_{mtl} = 0$.

Table V shows the target recognition performance evaluated by both the SC and DC on $D_{17,GAN}^{AC}$ and $D_{17,GAN}^{C}$ in comparison with $D_{17,GAN}^{Peace}$. As shown in Table V, the target recognition performances on $D_{17,GAN}^{Peace}$ by both SC and DC are always higher than those on $D_{17,GAN}^{AC}$ and $D_{17,GAN}^{C}$, which implies that the PeaceGAN can generate better SAR target images than the $ACGAN_m$ and $CGAN_m$ in terms of both the 'adequacy as training data' and the fidelity of 'target-class features', respectively. Fig 7 shows some of generated SAR target images by the $ACGAN_m$. As shown in Fig. 7, the shapes of the targets generally appear dim, and sometimes speckle noises look dissimilar to $D_{17}$. From this observation in Fig. 7 and the results in Table V, it can be noticed that the proposed PeaceGAN can not only control pose angles to generate SAR target images but also learn the joint distributions of SAR target images more reliably for various target classes and pose angles.

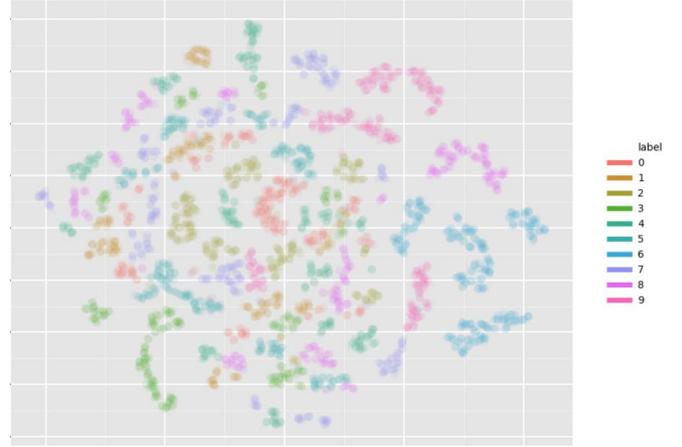

(a) 2D t-SNE of $D_{17,GAN}^{Peace}$

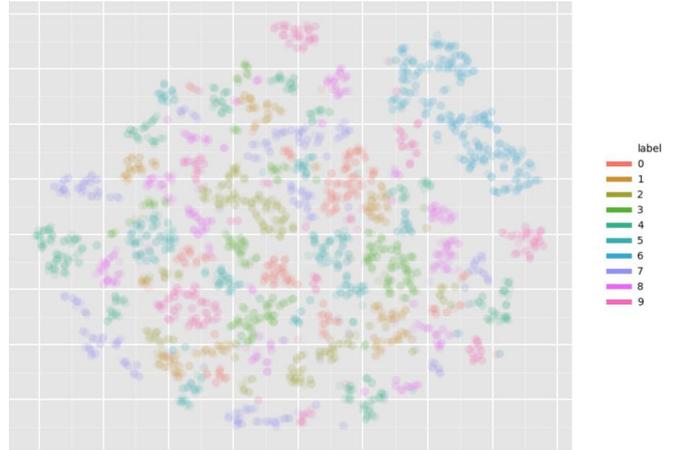

(b) 2D t-SNE of $D_{17,GAN}^{AC}$

Fig. 8. The comparison between 2D t-SNE of $D_{17,GAN}^{Peace}$ and $D_{17,GAN}^{AC}$.

We also use 2D t-SNE [45] to visualize the distributions of generated SAR target images of $D_{17,GAN}^{Peace}$ and $D_{17,GAN}^{AC}$ in a 2-dimensional space. Fig. 8 shows the distribution visualizations of 10 kinds of feature points for 64×64-sized 2,747 generated SAR target images (data samples) using the 2D t-SNE, where the data samples in a same class are represented by a same color. As shown in Fig. 8, the data samples in $D_{17,GAN}^{AC}$ are spread with larger overlaps across different classes than those of $D_{17,GAN}^{Peace}$. In other words, the data samples of

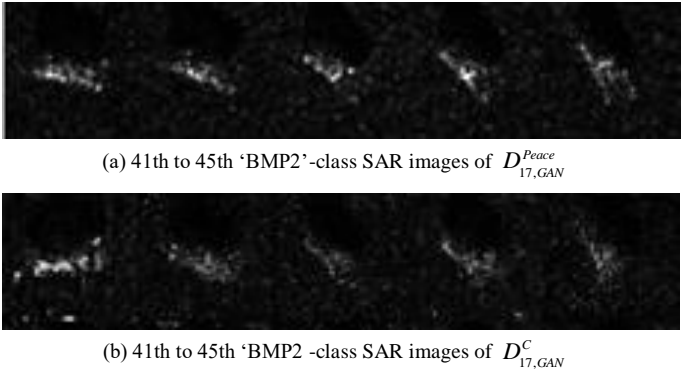

(a) 41th to 45th 'BMP2'-class SAR images of $D_{17,GAN}^{Peace}$

(b) 41th to 45th 'BMP2-class SAR images of $D_{17,GAN}^{C}$

Fig. 9. The comparison between examples of $D_{17,GAN}^{Peace}$ and $D_{17,GAN}^{C}$.

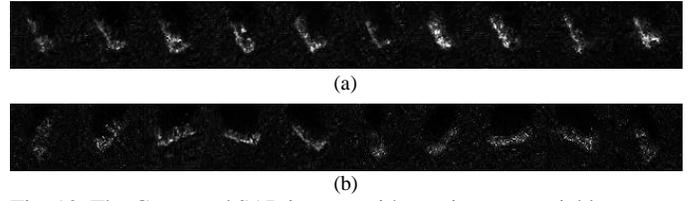

(a)

(b)

Fig. 10. The Generated SAR images with varying two variables, target classes and pose angles: (a) Fixed pose angle 151.2° and varying 10 target classes, (b) Fixed target class 'BTR70' and varying 10 pose angles from 0° to 324° uniformly by a pose angle increment of 36°.

TABLE VI
POSE ESTIMATION PERFORMANCE OF THE PRE-TRAINED DP-NET ON $D_{17,GAN}^{Peace}$ AND $D_{17,GAN}^{C}$.

| Classes | $D_{17,GAN}^{Peace}$ | | $D_{17,GAN}^{C}$ | |
|---|---|---|---|---|
| | MAD | STD | MAD | STD |
| BMP2 | 7.03 | 11.49 | 11.59 | 19.28 |
| BTR70 | 8.48 | 12.04 | 10.95 | 18.07 |
| T72 | 6.71 | 10.44 | 11.59 | 19.67 |
| BTR60 | 6.71 | 9.55 | 9.50 | 16.05 |
| 2S1 | 6.84 | 10.29 | 9.67 | 16.99 |
| BRDM2 | 6.77 | 10.73 | 10.61 | 18.38 |
| D7 | 7.45 | 10.38 | 10.61 | 16.41 |
| T62 | 6.39 | 10.27 | 9.61 | 16.98 |
| ZIL131 | 5.83 | 8.49 | 8.85 | 14.23 |
| ZSU234 | 6.10 | 8.99 | 8.72 | 13.32 |
| Averages | **6.83** | **10.27** | **10.17** | **16.94** |

$D_{17,GAN}^{Peace}$ belonging to the same classes exhibit higher concentrations than those of $D_{17,GAN}^{AC}$. Based on these observations, it can also be concluded that taking an advantage of pose angle information for SAR target image generation can help generate SAR target images more reliably with higher fidelities for various target classes.

On the other hand, Fig. 9 shows the examples of $D_{17,GAN}^{Peace}$ generated by the PeaceGAN and $D_{17,GAN}^{C}$ generated by the CGAN$_m$. As shown in Fig. 9, it also can be easily noted that the PeaceGAN can produce SAR target images visually better than CGAN$_m$. Furthermore, the generated SAR target images of $D_{17,GAN}^{C}$ tends to contain unnatural speckle noises and abnormal scattering characteristics. Furthermore, the DC result of CGAN$_m$ (95.63%) is slightly lower than that of ACGAN$_m$ (96.23%) as shown in Table V, which implies that the fidelity of 'target-class features' of ACGAN$_m$ is better than that of CGAN$_m$. On the other hand, please note that CGAN$_m$ can also control the pose angles to generate $D_{17,GAN}^{C}$, so it is reasonable that the $D_{17,GAN}^{C}$ has more diverse pose angles than $D_{17,GAN}^{AC}$ so the SC result of CGAN$_m$ (77.53%) is better than that of ACGAN$_m$ (73.03%) the perspective of 'adequacy as training data' (SC). Lastly, Table VI shows the pose estimation performance on $D_{17,GAN}^{Peace}$ and $D_{17,GAN}^{C}$ evaluated by DP to compare the fidelity of 'pose-angle features'. As shown in Table VI, the DP result of both MAD and STD of $D_{17,GAN}^{Peace}$ are much better than those of $D_{17,GAN}^{C}$. As a result, the proposed structure of PeaceGAN including pose estimator and auxiliary classifier also outperforms the CGAN-based structure from both quantitative and qualitative manners.

### D. SAR Image Generation with Varying Target Classes and Pose Angles

Fig. 10 shows the ability of PeaceGAN, as a generator, that can generate SAR target images for given desired pose angles and target classes. The SAR target images in Fig. 10-(a) were obtained by inputting into the PeaceGAN generator the target class information from 'BMP2' to 'ZSU234' sequentially while fixing the pose angle at 151.2°. Fig. 10-(b) shows the generated SAR target images of 'BTR70'-class for varying pose angles from 0° to 324° uniformly by a pose angle increment of 36°. As shown in Fig. 10, it is clear that the PeaceGAN can faithfully generate SAR target images with two controllable variables such as pose angles and target classes.

## VI. CONCLUSION

In this paper, we propose a novel GAN structure that can be jointly trained in an end-to-end manner with multi-task learning of estimating both pose angle and target class information of SAR target images, called PeaceGAN. As a result, the PeaceGAN explicitly disentangles the pose angles and the target class, which leads to both the diversity and the improved quality of the generated SAR target images that are visually very similar to real SAR target images. We also propose three indirect evaluation methods to show the adequacy of generated SAR target images as training data and the fidelities of target-class features and pose-angle features of them generated by our PeaceGAN. From intensive experiments, the PeaceGAN has shown a high potential to generate SAR target images with good fidelity at intended pose angles for desired target classes.